\documentclass[letterpaper, 10 pt, journal, twoside]{IEEEtran}

\IEEEoverridecommandlockouts                              

\title{GRAM: \\Generalization in Deep RL with a Robust Adaptation Module}

\author{James Queeney\textsuperscript{$1,2$}, Xiaoyi Cai\textsuperscript{$2$}, Alexander Schperberg\textsuperscript{$1$}, Radu Corcodel\textsuperscript{$1$},\\ Mouhacine Benosman\textsuperscript{$3$}, and Jonathan P.~How\textsuperscript{$2$}
\thanks{Supplementary Video: \url{https://youtu.be/0g2nos2EnCw}.}%
\thanks{Code: \url{https://github.com/merlresearch/gram}.}%
\thanks{Accepted for publication in IEEE Robotics and Automation Letters. JQ, AS, and RC were exclusively supported by MERL. XC and JPH were supported by ARL grant W911NF-21-2-0150 and by ONR grant N00014-18-1-2832. The work of JQ and MB was completed prior to joining Amazon Robotics. MB and JPH contributed equally as senior authors.}%
\thanks{\textsuperscript{$1$ }Mitsubishi Electric Research Laboratories (MERL), Cambridge, MA 02139, USA. {\tt\footnotesize \{queeney, schperberg, corcodel\}@merl.com}.}%
\thanks{\textsuperscript{$2$ }Massachusetts Institute of Technology, Cambridge, MA 02139, USA. {\tt\footnotesize \{queeney, xyc, jhow\}@mit.edu}.}%
\thanks{\textsuperscript{$3$ }Amazon Robotics, North Reading, MA 01864, USA. {\tt\footnotesize m\_benosman@ieee.org}.}%
}

\usepackage[hidelinks]{hyperref}
\usepackage{url}

\usepackage{amsmath,amssymb}
\usepackage[pdftex]{graphicx}
\usepackage{array}
\usepackage{booktabs}
\usepackage{colortbl}
\usepackage[sort,compress]{cite}

\usepackage[ruled,noline, noalgohanging]{algorithm2e}
\SetAlgorithmName{Alg.}{List of Algorithms}

\makeatletter
\newcommand{\removelatexerror}{\let\@latex@error\@gobble}
\makeatother

\newcommand{\figref}[1]{Fig.~\ref{#1}}
\newcommand{\tabref}[1]{Table~\ref{#1}}
\newcommand{\secref}[1]{Section~\ref{#1}}
\newcommand{\algref}[1]{Alg.~\ref{#1}}

\newcommand{\E}{\mathop{\mathbb{E}}}

\newcolumntype{C}[1]{>{\centering\arraybackslash}b{\dimexpr #1\linewidth}}
\newcolumntype{L}[1]{>{\raggedright\arraybackslash}b{\dimexpr #1\linewidth}}
\newcolumntype{R}[1]{>{\raggedleft\arraybackslash}b{\dimexpr #1\linewidth}}
\newcolumntype{M}[2]{>{\centering\arraybackslash}b{\dimexpr #1\linewidth+#2\tabcolsep}}

\begin{document}

\maketitle


\begin{abstract}
The reliable deployment of deep reinforcement learning in real-world settings requires the ability to generalize across a variety of conditions, including both in-distribution scenarios seen during training as well as novel out-of-distribution scenarios. In this work, we present a framework for dynamics generalization in deep reinforcement learning that unifies these two distinct types of generalization within a single architecture. We introduce a robust adaptation module that provides a mechanism for identifying and reacting to both in-distribution and out-of-distribution environment dynamics, along with a joint training pipeline that combines the goals of in-distribution adaptation and out-of-distribution robustness. Our algorithm GRAM achieves strong generalization performance across in-distribution and out-of-distribution scenarios upon deployment, which we demonstrate through extensive simulation and hardware locomotion experiments on a quadruped robot.
\end{abstract}



\section{Introduction}

\IEEEPARstart{D}{ue} to the diverse and uncertain nature of real-world settings, generalization is an important capability for the reliable deployment of data-driven, learning-based frameworks such as deep reinforcement learning (RL). Policies trained with deep RL must be capable of generalizing to a variety of different environment dynamics at deployment time, including both familiar training conditions and novel unseen scenarios, as the complex nature of real-world environments makes it difficult to capture all possible variations during training.


\begin{figure}[t]
    \centering
    \includegraphics[width=1.00\linewidth]{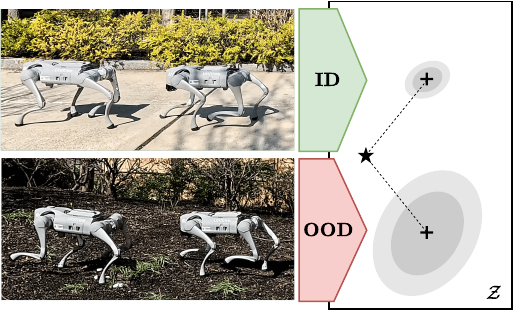}
    \vspace{-1.5em}
    \caption{GRAM generalizes to both ID and OOD environment dynamics at deployment time with a single unified architecture. GRAM introduces a robust adaptation module that quantifies uncertainty about the deployment environment using a recent history of observations, and biases latent context estimates towards a special robust latent feature ($\star$) when uncertainty is high.}
    \label{fig:overview}
    \vspace{-1.5em}
\end{figure}


For complex robotics applications such as legged locomotion, existing deep RL approaches to zero-shot dynamics generalization have focused on two complementary concepts: adaptation and robustness. Contextual RL techniques~\cite{hallak_2015} such as teacher-student training~\cite{lee_j_2020_quadruped,kumar_2021,margolis_2024} learn to identify and adapt to the current environment dynamics to achieve the best performance, but this adaptation is only reliable for the range of in-distribution (ID) scenarios seen during training. Robust RL methods~\cite{nilim_2005,iyengar_2005} such as adversarial training~\cite{shi_2024}, on the other hand, maximize the worst-case performance across a range of possible environment dynamics, providing generalization to out-of-distribution (OOD) scenarios at the cost of conservative performance in ID environments.

This work shows how to extract the benefits of these complementary approaches in a unified framework  called GRAM: \underline{G}eneralization in deep RL with a \underline{R}obust \underline{A}daptation \underline{M}odule. GRAM achieves both adaptive ID and robust OOD dynamics generalization at deployment time within a single architecture. Our main contributions are as follows:
\begin{enumerate}
    \item We introduce a \emph{robust adaptation module} in \secref{sec:ra_module} that provides a mechanism for identifying both ID and OOD environment dynamics within the same architecture. We extend existing contextual RL approaches by using an epistemic neural network~\cite{osband_2023} to incorporate a measure of uncertainty about the environment at deployment time, as illustrated in \figref{fig:overview}.
    \item We propose a joint training pipeline in \secref{sec:ra_train} that combines a teacher-student architecture for learning adaptive ID performance with adversarial RL training for robust OOD performance, resulting in a single unified policy that can achieve both ID and OOD dynamics generalization.
    \item We demonstrate the strong ID and OOD performance of GRAM through comprehensive locomotion experiments on a Unitree Go2 quadruped robot, including both simulation analysis in \secref{sec:experiments} and hardware results in \secref{sec:hardware}.
\end{enumerate}


\section{Related Work}

Deep RL has demonstrated success on complex robotics tasks in recent years, where policies are trained in simulation and deployed in the real world. In robotic control applications such as legged locomotion, it is critical for these learned policies to achieve zero-shot generalization across a range of scenarios at deployment time~\cite{kirk_2023}. In order to accomplish this, deep RL methods for robotic control have applied techniques from contextual RL~\cite{hallak_2015} and robust RL~\cite{nilim_2005,iyengar_2005}.

Contextual RL focuses on adapting across a range of ID environments seen during training, and many studies have demonstrated the importance of leveraging contextual information in deep RL to improve generalization (e.g., \cite{benjamins_2023}). In robotic control applications, the context is typically available in simulation during training, but unknown at deployment time where it must be inferred from past observations. The most common implementation of contextual RL for legged locomotion leverages privileged context information during training, and applies a teacher-student architecture to train a policy that can be deployed using only the history of past observations~\cite{lee_j_2020_quadruped,kumar_2021,margolis_2024}. Self-supervised techniques have also been applied to infer context from history in legged locomotion through the use of variational inference~\cite{lyu_2024} and contrastive learning objectives~\cite{long_2024}, and have been explored more broadly in the deep RL literature~\cite{yang_2020, chen_2022, ren_2022, lee_k_2020_context, luo_f_2022, nagabandi_2019, rakelly_2019, zintgraf_2020}. All of these methods are designed to adapt across the range of ID contexts seen during training, but are not specifically trained to handle OOD contexts with different environment dynamics. As a result, their ability to generalize is sensitive to the distribution of ID training contexts, and they may not generalize well to OOD contexts not considered during training.

Robust RL focuses on generalizing to OOD environments at deployment time by maximizing worst-case performance over a set of transition models~\cite{nilim_2005,iyengar_2005}. Deep RL methods for robotic control most commonly incorporate robustness through the use of adversarial training, which applies worst-case perturbations during training to provide robustness to unknown dynamics or disturbances at deployment time. Adversarial interventions in legged locomotion have primarily applied external forces to the robot~\cite{shi_2024, xiao_2024}, while more general adversarial RL methods have considered perturbations to actions~\cite{tessler_2019}, observations~\cite{zhang_2020}, and transitions~\cite{queeney_2023, zhang_2023, queeney_2024}. These approaches provide robust generalization to environment dynamics that were not explicitly seen during training, but often sacrifice ID performance to achieve robustness.

In this work, we are interested in both adaptive ID and robust OOD generalization. The possibility of different modes at deployment is related to deep RL methods that train a collection of policies to select from at deployment time~\cite{thananjeyan_2021, wagener_2021, ajay_2022, margolis_2023, he_2024, chen_2025}. This has been applied to legged locomotion by switching between a task-based policy and a recovery policy in order to guarantee safety~\cite{he_2024}, as well as selecting from a finite collection of different behaviors based on the deployment environment~\cite{margolis_2023, chen_2025}. Contextual RL can also be viewed as learning a collection of policies for ID adaptation, and our robust adaptation module extends this approach to incorporate a mode for robust OOD generalization as well.


\section{Problem Formulation}


\paragraph{Contextual RL} 

We model the problem of dynamics generalization in deep RL as a Contextual Markov Decision Process (CMDP)~\cite{hallak_2015}. A CMDP considers a set of contexts $\mathcal{C}$ that define a collection of MDPs $\left\lbrace \mathcal{M}_c \right\rbrace_{c \in \mathcal{C}}$. For each $c \in \mathcal{C}$, we have an MDP given by the tuple $\mathcal{M}_c = \left( \mathcal{S}, \mathcal{A}, p_c, r, \rho_0, \gamma \right)$, where $\mathcal{S}$ is the set of states, $\mathcal{A}$ is the set of actions, $p_c: \mathcal{S} \times \mathcal{A} \times \mathcal{C} \rightarrow P(\mathcal{S})$ is the context-dependent transition model where $P(\mathcal{S})$ represents the space of probability measures over $\mathcal{S}$, $r: \mathcal{S} \times \mathcal{A} \rightarrow \mathbb{R}$ is the reward function, $\rho_0$ is the initial state distribution, and $\gamma$ is the discount rate. We focus on the setting where the transition model $p_c$ depends on the context $c \in \mathcal{C}$ (i.e., varying dynamics), while the reward function $r$ remains the same across contexts (i.e., same task). For a policy $\pi$ and context $c \in \mathcal{C}$, performance is given by the expected total discounted returns 
\begin{equation}
J(\pi, c) = \E_{\tau \sim (\pi,c)} \left[ \sum_{t=0}^\infty \gamma^t r(s_t, a_t) \right],
\end{equation}
where $\tau = (s_0, a_0, s_1, \ldots)$ and $\tau \sim (\pi,c)$ represents a trajectory sampled by deploying the policy $\pi$ in the MDP~$\mathcal{M}_c$.


\paragraph{Problem statement} 

We assume that the context is available as privileged information during training, but is not available for deployment. This is often the case when a policy is trained in simulation and deployed in the real world. We train across a range of contexts to achieve generalization, but it is typically not possible to consider all possible contexts due to unknown factors at deployment time. Instead, we assume access to a subset of ID training contexts $\mathcal{C}_{\textnormal{ID}} \subset \mathcal{C}$, and we write $c \sim \mathcal{C}_{\textnormal{ID}}$ to represent a sample from a training distribution over ID contexts. We define $\mathcal{C}_{\textnormal{OOD}} = \mathcal{C} \setminus \mathcal{C}_{\textnormal{ID}}$ as the set of OOD contexts that are not seen during training. \emph{Our goal is to train a single policy that performs well in both ID and OOD contexts at deployment time}: 
\begin{equation}
\max_{\pi} \, J(\pi, c) \quad \forall c \in \mathcal{C} = \mathcal{C}_{\textnormal{ID}} \cup \mathcal{C}_{\textnormal{OOD}}.
\end{equation}
Because we have access to ID contexts during training, we can achieve adaptive ID generalization by directly maximizing performance for each $c \in \mathcal{C}_{\textnormal{ID}}$. On the other hand, we do not have access to OOD contexts during training, so we instead seek to achieve robust OOD generalization for $c \in \mathcal{C}_{\textnormal{OOD}}$ by applying techniques from robust RL. Note that adaptive ID performance and robust OOD performance represent two distinct types of generalization with different objectives, making it challenging to achieve both with a single policy. 


\paragraph{Teacher-student training} 

The teacher-student approach to generalization in deep RL assumes access to $c_t \in \mathcal{C}_{\textnormal{ID}}$ at every timestep throughout training, and leverages this privileged context information to train a teacher policy that can adapt to different contexts. The teacher policy applies a context encoder $f: \mathcal{C} \rightarrow \mathcal{Z}$ that maps the context to a latent feature $z_t = f(c_t)$, which is then provided as an input to the policy $\pi: \mathcal{S} \times \mathcal{Z} \rightarrow P(\mathcal{A})$ and critic $V^{\pi}: \mathcal{S} \times \mathcal{Z} \rightarrow \mathbb{R}$. In this work, we consider the latent feature space $\mathcal{Z} = \mathbb{R}^d$. The context encoding $z_t = f(c_t)$, policy $\pi(a_t \mid s_t, z_t)$, and critic $V^\pi(s_t, z_t)$ are trained to minimize the average actor-critic RL loss over ID contexts given by
\begin{equation}\label{eq:rl_loss}
    \mathcal{L}_{\textnormal{RL}} = \E_{c \sim \mathcal{C}_{\textnormal{ID}}} \left[ \mathcal{L}_{\pi}(c) + \mathcal{L}_{V}(c) \right],
\end{equation}
where $\mathcal{L}_{\pi}(c)$ and $\mathcal{L}_{V}(c)$ represent the policy loss and critic loss, respectively, of a given RL algorithm for the context $c \sim \mathcal{C}_{\textnormal{ID}}$. In our experiments, we apply Proximal Policy Optimization (PPO) \cite{schulman_2017} as the RL algorithm.

Note that the teacher policy cannot be applied at deployment time because it requires privileged information about the context $c_t$ in order to compute the latent feature $z_t$. For this reason, RL training is followed by a supervised learning phase where a student policy is trained to imitate the teacher policy using only the recent history of states and actions from the last $H$ timesteps $h_t = \left(s_{t-H}, a_{t-H}, \ldots, s_t \right) \in \mathcal{H}$. In particular, an adaptation module $\phi: \mathcal{H} \rightarrow \mathcal{Z}$ that maps recent history to a latent feature $\hat{z}_t = \phi(h_t)$ is trained to minimize the loss
\begin{equation}
    \mathcal{L}_{\textnormal{enc}} = \E_{c \sim \mathcal{C}_{\textnormal{ID}}} \left[  \E_{\tau \sim (\pi,c)} \left[ \left\Vert f(c_t) - \phi(h_t) \right\Vert^2 \right] \right], \label{eq:enc_loss}
\end{equation}
where expectation is taken with respect to trajectories sampled using the student policy in ID contexts $c \sim \mathcal{C}_{\textnormal{ID}}$ during training. This training represents a form of implicit system identification across ID contexts. Using the history encoding $\hat{z}_t = \phi(h_t)$, the policy $\pi(a_t \mid s_t, \hat{z}_t)$ can be applied at deployment time because it does not require privileged information as input.


\section{Robust Adaptation Module}\label{sec:ra_module}

We build upon the teacher-student architecture for adaptation in deep RL, which has demonstrated strong performance in complex robotics applications such as legged locomotion~\cite{lee_j_2020_quadruped,kumar_2021,margolis_2024}. However, because this approach focuses only on adaptation across ID contexts seen during training, its OOD generalization capabilities depend strongly on the relationship between $\mathcal{C}_{\textnormal{ID}}$ and $\mathcal{C}_{\textnormal{OOD}}$. The learned adaptation module $\phi$ is trained to identify ID contexts from history, so its output $\hat{z}_t = \phi(h_t)$ and resulting policy $\pi(a_t \mid s_t, \hat{z}_t)$ are only reliable for the distribution of ID history inputs $h_t$ that were observed during training. This leads to strong performance across $c \in \mathcal{C}_{\textnormal{ID}}$, but the standard adaptation module estimate $\hat{z}_t = \phi(h_t)$ may not be useful for achieving generalization in OOD contexts $c \in \mathcal{C}_{\textnormal{OOD}}$ with different environment dynamics not seen during training.

\begin{figure}[t]
    \centering
    \includegraphics[width=\linewidth]{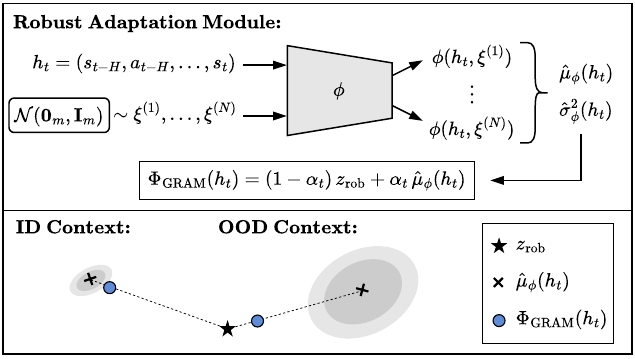}
    \vspace{-1.0em}
    \caption{Robust adaptation module used by GRAM at deployment time. Top: Epistemic neural network $\phi$ outputs a sample mean and variance of latent feature estimates for a history $h_t$, which are used to calculate $\Phi_{\textnormal{GRAM}}$ in \eqref{eq:ra_module}. Bottom: In ID contexts, variance of latent feature estimates will be low and $\Phi_{\textnormal{GRAM}}$ will be close to the mean estimate. In OOD contexts with different environment dynamics, variance will be high and $\Phi_{\textnormal{GRAM}}$ will output an estimate close to $z_{\textnormal{rob}}$.}
    \label{fig:ram}
    \vspace{-1.0em}
\end{figure}

In order to generalize to both ID and OOD contexts at deployment time within the same architecture, we introduce a \emph{robust adaptation module} that explicitly incorporates a mechanism for identifying and reacting to OOD contexts. We accomplish this by quantifying the level of uncertainty present in the latent feature estimate $\hat{z}_t$. We represent the adaptation network $\phi$ as an epistemic neural network \cite{osband_2023} with the form
\begin{equation}\label{eq:epinet}
    \phi(h_t, \xi) = \phi_{\textnormal{base}}(h_t) + \phi_{\textnormal{epi}} ( \tilde{h}_t, \xi ),
\end{equation}
where $\xi \in \mathbb{R}^m$ is a random input that is sampled from a multivariate standard Gaussian distribution $q = \mathcal{N}(\mathbf{0}_m,\mathbf{I}_m)$, and $\tilde{h}_t$ is the concatenation of $h_t$ and the output from the last hidden layer of $\phi_{\textnormal{base}}$ with gradients stopped as in \cite{osband_2023}. By incorporating a random input in the second component of \eqref{eq:epinet} (i.e., the ``epinet''), this architecture provides a distribution of latent feature estimates for a history input $h_t$ rather than a single point estimate, as illustrated on the top of \figref{fig:ram}. For a history $h_t$ and $N$ random input samples $\xi^{(1)},\ldots,\xi^{(N)} \sim q$, we can write the sample mean and variance of the latent feature estimates as
\begin{align}
    \hat{\mu}_{\phi}(h_t) &= \frac{1}{N} \sum_{i=1}^N \phi(h_t, \xi^{(i)}), \label{eq:epinet_mean} \\ 
    \hat{\sigma}^2_{\phi}(h_t) &=  \frac{1}{N-1} \sum_{i=1}^N ( \phi(h_t, \xi^{(i)}) - \hat{\mu}_{\phi}(h_t)  )^2, \label{eq:epinet_var}
\end{align}
where all operations are performed per-dimension. We train~\eqref{eq:epinet} to minimize the encoder loss across random input samples $\xi \sim q$, resulting in the modified encoder loss given by 
\begin{equation}\label{eq:epi_enc_loss}
    \mathcal{L}^{\textnormal{GRAM}}_{\textnormal{enc}} = \E_{\xi \sim q} \left[ \E_{c \sim \mathcal{C}_{\textnormal{ID}}} \left[  \E_{\tau \sim (\pi,c)} \left[ \left\Vert f(c_t) - \phi(h_t, \xi) \right\Vert^2 \right] \right]  \right].
\end{equation}
By doing so, the total variance of the latent feature estimates $\Vert \hat{\sigma}_{\phi}(h_t) \Vert^2$ will be small over the distribution of history inputs $h_t$ that were seen during training (i.e., trajectories sampled from ID contexts), but not for histories in OOD contexts with different dynamics not seen during training.


Using the epistemic neural network architecture in \eqref{eq:epinet}, we introduce a \emph{robust adaptation module} to generalize to both ID and OOD contexts at deployment time. When uncertainty of the latent feature estimates is low, we output the mean estimate $\hat{\mu}_{\phi}(h_t)$ to allow for adaptation in ID contexts. When uncertainty of the latent feature estimates is high, we bias the mean estimate towards a special robust latent feature $z_{\textnormal{rob}} = \mathbf{0}_d$ to identify that OOD dynamics have been detected. For a given history $h_t$, our robust adaptation module $\Phi_{\textnormal{GRAM}}: \mathcal{H} \rightarrow \mathcal{Z}$ outputs a latent feature $\hat{z}_t = \Phi_{\textnormal{GRAM}}(h_t)$ according to
\begin{equation}\label{eq:ra_module}
\begin{gathered}
    \Phi_{\textnormal{GRAM}}(h_t) = (1-\alpha_t) \, z_{\textnormal{rob}} + \alpha_t \, \hat{\mu}_{\phi}(h_t), \\
    \alpha_t = \exp \left( -\beta ( \Vert \hat{\sigma}_{\phi}(h_t) \Vert^2 - \delta )_+ \right),
\end{gathered}
\end{equation}
where $(\, \cdot \,)_+ = \max(\, \cdot \, , 0)$ and $\alpha_t \in [0,1]$, with $\alpha_t \rightarrow 1$ when uncertainty is low and $\alpha_t \rightarrow 0$ when uncertainty is high. This formulation is motivated by the posterior predictive mean in evidential deep learning~\cite{ulmer_2023}, where $z_{\textnormal{rob}}$ represents a robust prior, $\hat{\mu}_{\phi}(h_t)$ is a data-driven estimate, and $\alpha_t$ is the normalized evidence. We include a scale parameter $\beta$ and shift parameter $\delta$ to allow for easy finetuning of $\alpha_t$ based on the output magnitude of the trained epinet in ID contexts, which can be calculated using a validation set of $\Vert \hat{\sigma}_{\phi}(h_t) \Vert^2$ values collected at the end of training. 

By defining a special robust latent feature $z_{\textnormal{rob}}$, we incorporate the failure mode of OOD dynamics directly into the existing adaptation framework as a single instance in latent feature space. This allows us to leverage the existing teacher-student training procedure for adaptation in ID contexts, while applying tools from robust RL to encode robust behavior into $\pi(a_t \mid s_t, z_{\textnormal{rob}})$ for OOD generalization \emph{within the same architecture}. Note that the privileged context encoding $z_t = f(c_t)$ begins training near $z_{\textnormal{rob}} = \mathbf{0}_d$ for a randomly initialized context encoder, and the RL loss $\mathcal{L}_{\textnormal{RL}}$ in \eqref{eq:rl_loss} incentivizes $z_t$ to move away from $z_{\textnormal{rob}}$ as needed to achieve adaptive performance. At deployment time, we bias the latent feature estimate back towards the robust anchor point $z_{\textnormal{rob}}$ if the estimate is unreliable due to OOD environment dynamics. See the bottom of \figref{fig:ram} for an illustration.


\section{Training for Robust Adaptation}\label{sec:ra_train}

\begin{figure}[t]
    \centering
    \vspace{-1.0em}
    \includegraphics[width=\linewidth]{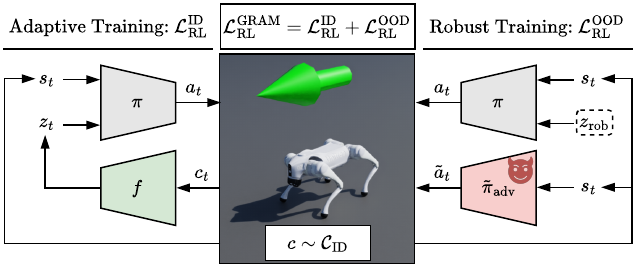}
    \vspace{-1.0em}
    \caption{Joint RL training pipeline used by GRAM, which combines standard ID data collection and adversarial data collection for every RL update. Training environments are assigned to \emph{adaptive training} or \emph{robust training} at each iteration, and assignments alternate between iterations. RL training is followed by supervised learning to train the adaptation network $\phi(h_t, \xi)$.}
    \label{fig:rl_training}
    \vspace{-1.0em}
\end{figure}

Our robust adaptation module provides an intuitive structure for achieving both ID and OOD dynamics generalization within a single architecture. In order to accomplish this goal, we jointly train our policy $\pi(a_t \mid s_t, z_t)$ for adaptive performance in ID environments and robust performance in OOD environments (i.e., when $z_t = z_{\textnormal{rob}}$). We consider parallel training environments as in \cite{rudin_2022}, and assign each training environment to either \emph{adaptive training} or \emph{robust training}. This assignment determines how the latent feature vector is calculated during training, as well as how data collection occurs in the environment. See \algref{alg:rl_training} and \figref{fig:rl_training} for an overview of the joint RL training pipeline, which is followed by a supervised learning phase to train the adaptation network $\phi(h_t, \xi)$. Note that the robust adaptation module $\Phi_{\textnormal{GRAM}}(h_t)$ is only applied at deployment time and not during training.


\paragraph{RL training}

Within each iteration of RL training, all data for a given training environment is collected according to either standard ID data collection or adversarial data collection, as described in the following. This provides temporal consistency when training the policy for adaptive or robust performance, respectively. We alternate these assignments between iterations, which allows full trajectories to contain a mixture of both forms of data. As shown in our experiments, this mixed data collection design provides additional robustness benefits compared to using the same training assignment for entire trajectories.

For environments assigned to \emph{adaptive training}, we follow the same data collection and training updates as the standard teacher-student architecture. We calculate the privileged latent feature vector as $z_t = f(c_t)$, and perform standard data collection with $\pi(a_t \mid s_t, z_t)$. We update the policy~$\pi$, critic~$V^{\pi}$, and context encoder~$f$ according to \eqref{eq:rl_loss}, and we denote this loss by $\mathcal{L}^{\textnormal{ID}}_{\textnormal{RL}}$.

For environments assigned to \emph{robust training}, we use the robust latent feature vector $z_{\textnormal{rob}}$ and apply an adversarial RL training pipeline to provide robustness to worst-case environment dynamics at deployment time.  We perform data collection with $\pi(a_t \mid s_t, z_{\textnormal{rob}})$, and we introduce an adversary policy $\tilde{\pi}_{\textnormal{adv}}: \mathcal{S} \rightarrow P(\Tilde{\mathcal{A}}_{\textnormal{adv}})$ that is trained with RL to minimize the returns of $\pi(a_t \mid s_t, z_{\textnormal{rob}})$. In our experiments on a Unitree Go2 quadruped robot, the adversary applies forward and lateral external forces to the robot's body 5\% of the time during data collection, where the direction of the external force is learned by the adversary. Using this adversarially collected data, we update the policy $\pi$, critic $V^{\pi}$, and adversary policy $\tilde{\pi}_{\textnormal{adv}}$ according to \eqref{eq:rl_loss} with the robust latent feature vector $z_{\textnormal{rob}}$ as input. We denote this loss by $\mathcal{L}^{\textnormal{OOD}}_{\textnormal{RL}}$. Note that adversarial RL provides a method for sampling worst-case trajectories during training, which leads to robust generalization in unseen OOD contexts at deployment time when the robust latent feature $z_{\textnormal{rob}}$ is provided to the policy as input. 

By combining \emph{adaptive training} and \emph{robust training}, the RL loss used by GRAM is given by
\begin{equation}\label{eq:gram_rl_loss}
\mathcal{L}^{\textnormal{GRAM}}_{\textnormal{RL}} = \mathcal{L}^{\textnormal{ID}}_{\textnormal{RL}} + \mathcal{L}^{\textnormal{OOD}}_{\textnormal{RL}}.
\end{equation}


\begin{figure}[t]
\removelatexerror

\SetAlgoSkip{medskip}
\SetAlCapNameFnt{\small}
\SetAlCapFnt{\small}

\begin{algorithm}[H]
\small
\caption{Joint RL training pipeline}\label{alg:rl_training}

\For{$K$ updates}{
    \BlankLine
    In \emph{adaptive training} environments, collect standard ID data with $\pi(a_t \mid s_t, z_t), \, z_t = f(c_t)$. Calculate $\mathcal{L}^{\textnormal{ID}}_{\textnormal{RL}}$.
    \BlankLine
    In \emph{robust training} environments, collect adversarial data with $\pi(a_t \mid s_t, z_{\textnormal{rob}})$ and $\tilde{\pi}_{\textnormal{adv}}(\tilde{a}_t \mid s_t)$. Calculate $\mathcal{L}^{\textnormal{OOD}}_{\textnormal{RL}}$.
    \BlankLine
    Optimize the RL loss $\mathcal{L}^{\textnormal{GRAM}}_{\textnormal{RL}} = \mathcal{L}^{\textnormal{ID}}_{\textnormal{RL}} + \mathcal{L}^{\textnormal{OOD}}_{\textnormal{RL}}$.
    \BlankLine
    Alternate training environment assignments.
	\BlankLine
}
\end{algorithm}

\end{figure}



\paragraph{Adaptation module training}

As in the standard teacher-student architecture, RL training is followed by a supervised learning phase to train the adaptation module for deployment using data collected with the student policy. We accomplish this by training the epistemic neural network architecture $\phi(h_t, \xi)$ in \eqref{eq:epinet} on the modified encoder loss $\mathcal{L}^{\textnormal{GRAM}}_{\textnormal{enc}}$ in \eqref{eq:epi_enc_loss}. Data collection uses the student policy $\pi(a_t \mid s_t, \hat{z}_t)$ with $\hat{z}_t = \hat{\mu}_{\phi}(h_t)$ or $\hat{z}_t = z_{\textnormal{rob}}$, depending on the training assignment. We do not apply an adversary during the supervised learning phase, as the goal is to train the epinet in \eqref{eq:epinet} to output estimates with low variance in ID contexts. 


\section{GRAM Algorithm}

Together, the robust adaptation module in \eqref{eq:ra_module} and the training procedure in \secref{sec:ra_train} form our algorithm GRAM. GRAM combines standard ID data collection and adversarial data collection during training to optimize the RL loss $\mathcal{L}^{\textnormal{GRAM}}_{\textnormal{RL}}$, followed by a supervised learning phase to optimize the encoder loss $\mathcal{L}^{\textnormal{GRAM}}_{\textnormal{enc}}$. Finally, by applying the robust adaptation module $\Phi_{\textnormal{GRAM}}$ at deployment time, our policy achieves both adaptive ID and robust OOD dynamics generalization within a single unified architecture.


\paragraph{Implementation details} 

In our experiments, we apply PPO \cite{schulman_2017} as the base RL algorithm with the parallel training scheme from \cite{rudin_2022}. We perform $10{,}000$ updates during RL training, followed by a supervised learning phase where we train the adaptation network for $5{,}000$ updates. We consider feedforward neural networks for all trainable components as in \cite{margolis_2024}, and we follow the design choices proposed in \cite{osband_2023} to model the epinet. The adversary policy is trained to apply worst-case external forces that cause abrupt changes to the robot's forward and lateral linear velocity. The direction of the external force is learned by the adversary, and the impact of the force on linear velocity increases from $0.0$ to $1.0$ meters per second over the course of training. See \tabref{tab:implementation} and our code for additional implementation details.

GRAM uses the epinet architecture to quantify uncertainty \emph{relative} to the uncertainty estimates in ID contexts. We accomplish this by including a scale parameter~$\beta$ and shift parameter~$\delta$ in the calculation of $\alpha_t$ in \eqref{eq:ra_module}, which we finetune at the end of training using a validation set of $\Vert \hat{\sigma}_{\phi}(h_t) \Vert^2$ values collected from ID contexts. We set $\delta$ to be the $u_{\textnormal{min}} = 0.90$ quantile of the validation set (note that $\alpha_t = 1.00$ when $\Vert \hat{\sigma}_{\phi}(h_t) \Vert^2 \leq \delta$), and we set $\beta$ such that $\alpha_t = 0.01$ at the $u_{\textnormal{max}} = 0.99$ quantile of the validation set. 


\begin{table}[t]
\footnotesize
\caption{Implementation details}
\vspace{-0.25em}
\label{tab:implementation}
\centering
\begin{tabular}{lc}
\toprule
Hyperparameter & Value \\
\midrule
Network architectures: MLP hidden layers \\
\cmidrule(lr){1-1} 
Policy ($\pi$), critic ($V^\pi$), adversary policy ($\tilde{\pi}_{\textnormal{adv}}$) & 512, 256, 128 \\
Context encoder ($f$) & 64, 64 \\
Base adaptation network ($\phi_{\textnormal{base}}$) & 512, 256, 128  \\
Epinet ($\phi_{\textnormal{epi}}$) & 16, 16  \\\\
Robust adaptation module hyperparameters \\
\cmidrule(lr){1-1} 
Context encoder latent feature size ($d$) & 8 \\
History length ($H$) & 16 \\
Epinet random input dimension ($m$) & $8$ \\
Epinet random input samples per data point ($N$) & $8$ \\
\bottomrule
\end{tabular}
\vspace{-1.5em}
\end{table}


As we show in the following sections, we found that the use of a teacher-student architecture for ID adaptation and adversarial RL for OOD robustness result in strong performance on the quadruped robot locomotion tasks we consider in our experiments. However, note that it is also possible to apply GRAM with different choices of contextual RL methods for ID adaptation and robust RL methods for OOD generalization, which represents an interesting avenue for future work.


\section{Simulation Experiments}\label{sec:experiments}

First, we evaluate the performance of GRAM on realistic simulated locomotion tasks with the Unitree Go2 quadruped robot in Isaac Lab \cite{mittal_2023}. The goal of the robot is to track a velocity command $\mathbf{v}_t^{\textnormal{cmd}} = [v^{\textnormal{cmd}}_{x}, v^{\textnormal{cmd}}_{y}, \omega^{\textnormal{cmd}}_{z}] \in \mathbb{R}^3$ provided as input, where $v^{\textnormal{cmd}}_x, v^{\textnormal{cmd}}_y$ represent target forward and lateral linear velocities, respectively, and $\omega^{\textnormal{cmd}}_z$ represents a target yaw angular velocity. For each episode, we uniformly sample linear velocity commands between $-1.0$ and $1.0$ meters per second (i.e., $v^{\textnormal{cmd}}_x, v^{\textnormal{cmd}}_y \sim \mathcal{U}([-1.0,1.0])$) and calculate the yaw angular velocity command $\omega^{\textnormal{cmd}}_z$ throughout the episode based on a target heading direction. The simulated quadruped robot can be seen in \figref{fig:rl_training}, with the velocity command represented by a green arrow.

The policy has access to noisy proprioceptive observations available from standard onboard sensors at every timestep (joint angles, joint velocities, projected gravity, and base angular velocities), and outputs target joint angles $a_t \in \mathbb{R}^{12}$ for each of the robot's 12 degrees of freedom that are converted to torques by a PD controller operating at 200~Hz with proportional gain $K_p = 25$ and derivative gain $K_d = 0.5$. The maximum episode length is 20 seconds with target joint angles processed at 50~Hz, which corresponds to 1{,}000 timesteps per episode. We consider the reward function used in \cite{margolis_2024}, which includes rewards for tracking the velocity command~$\mathbf{v}_t^{\textnormal{cmd}}$ and regularization terms to promote smooth and stable gaits. See~\cite{margolis_2024} for details.


\begin{table}[t]
\footnotesize
\caption{ID context set for training}
\vspace{-0.25em}
\label{tab:base_id}
\centering
\begin{tabular}{lccc}
\toprule
Parameter & Dim. & Nominal & Range \\
\midrule
Friction multiple & 1 & 1.00 & $\left[ 0.05, 4.50 \right]$ \\
Added base mass (kg) & 1 & 0.00 & $\left[ -1.00, 3.00 \right]$ \\
Motor strength multiple & 12 & 1.00 & $\left[ 0.80, 1.20 \right]$ \\
Joint angle bias (rad) & 12 & 0.00 & $\left[ -0.10, 0.10 \right]$ \\
\bottomrule
\end{tabular}
\vspace{-1.5em}
\end{table}



\paragraph{Benchmark comparison} 

We compare the performance of GRAM against popular deep RL methods for generalization in legged locomotion: robust RL using adversarial training \cite{shi_2024}, contextual RL using a teacher-student architecture~\cite{lee_j_2020_quadruped,kumar_2021,margolis_2024}, and domain randomization~\cite{peng_2018}. Our goal is to investigate the generalization behavior of deep RL methods when faced with deployment environments that differ from those used in training. We conduct this analysis by training all methods on flat, solid ground across the ID context set $\mathcal{C}_{{\textnormal{ID}}}$ described in \tabref{tab:base_id}, which represents moderate dynamics variations that are commonly used to promote sim-to-real transfer. We evaluate the average velocity-tracking task returns normalized to $[0,1]$ for all algorithms across a variety of deployment settings in \tabref{tab:sim_results}. We consider deployment in ID, near OOD, and far OOD contexts. Near OOD contexts consider moderate OOD variations to contexts sampled from $\mathcal{C}_{{\textnormal{ID}}}$ in \tabref{tab:sim_results}: (i)~6~kg added base mass, (ii)~ 5\textdegree{} incline, (iii)~5~cm rough terrain, and (iv)~frozen back hip joint. Far OOD contexts consider more difficult OOD variations: (i)~9~kg added base mass, (ii)~ 10\textdegree{} incline, (iii)~10~cm rough terrain, and (iv)~random frozen joint.


\begin{table}[t]
\footnotesize
\caption{Simulation results}
\vspace{-0.25em}
\label{tab:sim_results}
\centering
\begin{tabular}{L{0.24} *{3}{C{0.18}}}
\toprule
 & \multicolumn{3}{M{0.54}{2}}{\quad Average Normalized Task Returns~$\uparrow$} \\
\cmidrule(lr){2-4}
Algorithm & ID & Near OOD & Far OOD \\
\midrule
GRAM & $0.92 \pm 0.01$ & \cellcolor[gray]{.9}$\mathbf{0.79 \pm 0.02}$ & \cellcolor[gray]{.9}$\mathbf{0.58 \pm 0.01}$ \\
Robust RL & $0.87 \pm 0.01$ & $0.74 \pm 0.02$ & \cellcolor[gray]{.9}$\mathbf{0.59 \pm 0.01}$ \\
Contextual RL & \cellcolor[gray]{.9}$\mathbf{0.94 \pm 0.00}$ & $0.75 \pm 0.02$ & $0.42 \pm 0.03$ \\
Domain Rand. & $0.93 \pm 0.01$ & $0.74 \pm 0.03$ & $0.43 \pm 0.03$ \\[1.0em] 
GRAM Ablations \\
\cmidrule(l){1-1}
No $\Phi_{\textnormal{GRAM}}$ & $0.93 \pm 0.00$ & \cellcolor[gray]{.9}$\mathbf{0.80 \pm 0.01}$ & $0.55 \pm 0.02$ \\
Separate Data & $0.92 \pm 0.00$ & $0.74 \pm 0.02$ & $0.46 \pm 0.03$ \\
Modular & \cellcolor[gray]{.9}$\mathbf{0.94 \pm 0.00}$ & $0.74 \pm 0.02$ & $0.41 \pm 0.02$ \\
\bottomrule
\end{tabular}
\end{table}



\begin{figure}[t]
    \centering
    \vspace{-0.5em}
    \includegraphics[width=1.00\linewidth]{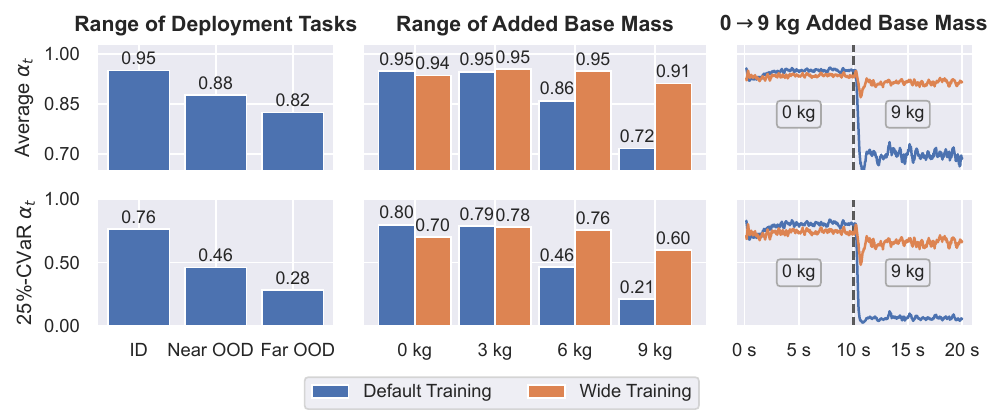}
    \vspace{-1.5em}
    \caption{GRAM average $\alpha_t$ (top) and 25\%-CVaR $\alpha_t$ (bottom). Left: Deployment environments from \tabref{tab:sim_results} with default ID context set for training. Middle: Range of added base mass deployment scenarios (default vs.~wide ID context sets for training). Right: 9 kg base mass added 10~seconds into deployment (default vs.~wide ID context sets for training).}
    \label{fig:gram_coefficient}
    \vspace{-1.3em}
\end{figure}


We see in \tabref{tab:sim_results} that contextual RL achieves the best ID performance as expected. GRAM achieves performance that is similar to contextual RL and domain randomization in ID contexts, while robust RL is overly conservative in this setting. Conservative ID performance is the main drawback of robust RL methods, which can be meaningful in practical real-world settings where we expect to encounter ID conditions the majority of the time. Importantly, GRAM does not encounter the same ID performance issues as robust RL. Instead, GRAM demonstrates adaptive performance by detecting low uncertainty in ID environments.

As we consider generalization beyond the ID context set seen during training, GRAM automatically identifies and reacts to the increased uncertainty of OOD deployment environments by decreasing the coefficient $\alpha_t$ in \eqref{eq:ra_module}. See \figref{fig:gram_coefficient} for the average and 25\% conditional value at risk (CVaR) of~$\alpha_t$ at deployment time. Note that OOD environments lead to high uncertainty at certain timesteps during deployment (resulting in low 25\%-CVaR $\alpha_t$), but for many timesteps the adaptive locomotion skills learned during training can still be applied (resulting in moderate $\alpha_t$ on average). GRAM leverages its adaptation capabilities to maintain strong performance in near OOD contexts without being overly conservative, outperforming all other methods in this setting. As deployment environments become more difficult in the far OOD setting, GRAM achieves robust performance comparable to robust RL. The non-robust baselines contextual RL and domain randomization, on the other hand, experience a more dramatic performance decline. Overall, existing baseline methods demonstrate trade-offs between ID and OOD performance, while GRAM can achieve strong ID and OOD generalization with a single unified policy.


\paragraph{Impact of ID context set for training}

\begin{figure}[t]
    \centering
    \includegraphics[width=1.00\linewidth]{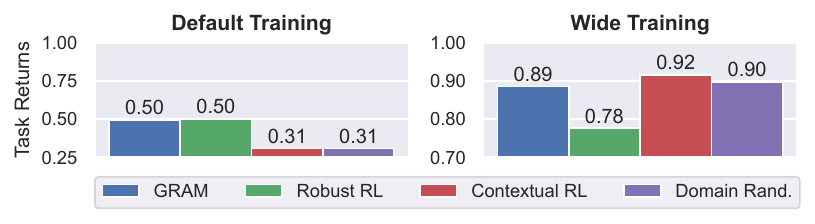}
    \vspace{-1.5em}
    \caption{Average normalized task returns for 9~kg added base mass scenario. Left: Training with default ID context set from \tabref{tab:base_id}. Right: Training across wide added base mass range of $[-1.00, 9.00]$.}
    \label{fig:id_training_impact}
    \vspace{-1.15em}
\end{figure}

Next, we analyze how the set of ID training contexts $\mathcal{C}_{{\textnormal{ID}}}$ impacts the performance of each algorithm. In addition to training with the default ID context set in \tabref{tab:base_id}, we also significantly expanded the range of added base masses seen during training to $[-1.00, 9.00]$. For both training setups, we compare performance of the learned policies at deployment time for a 9~kg added base mass in \figref{fig:id_training_impact}, which represents a far OOD scenario under the default training setup but an ID scenario when trained with the wide mass range.

We see in \figref{fig:id_training_impact} that the generalization capabilities of the baseline algorithms strongly depend on the set of ID contexts seen during training. When trained on the default range of added base masses, a 9~kg added base mass is far OOD and robust RL outperforms the non-robust baselines. When trained on a wider range of added base masses that includes 9~kg, contextual RL achieves the best performance by adapting to the ID deployment environment while robust RL becomes overly conservative. Unlike the baseline algorithms, GRAM identifies ID contexts from OOD contexts at deployment time in a way that automatically adjusts for different choices of~$\mathcal{C}_{{\textnormal{ID}}}$. As shown in \figref{fig:gram_coefficient}, GRAM detects low uncertainty (high average and 25\%-CVaR $\alpha_t$) across a range of added base masses when trained on the wide ID context set because these scenarios were all seen during training, resulting in adaptive performance similar to contextual RL. When trained on the default ID context set, GRAM decreases $\alpha_t$ for large added base masses to capture OOD environment uncertainty, achieving robust performance similar to robust RL.


\paragraph{Ablation analysis}


Finally, we conduct an ablation study in \tabref{tab:sim_results} to analyze the impact of GRAM's unified architecture and joint training pipeline. We compare GRAM to an ablation that applies our joint RL training pipeline without the robust adaptation module, a variant that applies separate ID and adversarial data collection to train a unified policy, and a modular approach that combines separate robust and adaptive policies at deployment time using an uncertainty threshold for switching. We see that these approaches can lead to comparable or slightly improved ID performance, but this comes at the cost of robustness in OOD scenarios. GRAM's single unified policy with its robust adaptation module and joint training pipeline  provide both adaptive ID and robust OOD performance, a combination that is difficult to achieve with other approaches.


\section{Hardware Experiments}\label{sec:hardware}

In addition to simulation experiments, we evaluate the real-world performance of GRAM by deploying our trained policies onboard a Unitree Go2 quadruped robot. We consider zero-shot sim-to-real transfer of the policies trained in simulation. All policies were trained on flat, solid ground with the moderate variations described by $\mathcal{C}_{{\textnormal{ID}}}$ in \tabref{tab:base_id}, which differs from prior works on legged locomotion that evaluate generalization after training on a broad set of diverse environments (e.g.,~\cite{kumar_2021}). As in simulation, the policy outputs target joint angles at 50 Hz, which are converted to torques by the built-in PD controller at 200 Hz with $K_p = 25$ and $K_d = 0.5$. Observations are available from standard onboard sensors, and all computation is performed onboard.


\paragraph{Controlled experiments} 

First, we considered controlled hardware experiments across 7 different scenarios, with the task of forward locomotion for $6$ meters from start to goal using a forward velocity command of $1.0$ meters per second (i.e., $v^{\textnormal{cmd}}_x = 1.0, v^{\textnormal{cmd}}_y = 0.0$). We measured the success rate and time-to-goal of GRAM, contextual RL, and robust RL across each scenario. We collected $5$ trials per scenario for each algorithm, resulting in a total of $105$ controlled hardware trials. 


\begin{table}[t]
\footnotesize
\caption{Hardware results: Moderate tasks}
\vspace{-0.25em}
\label{tab:hardware_results_moderate}
\centering
\begin{tabular}{L{0.23} *{3}{C{0.19}}}
\toprule
 & \multicolumn{3}{M{0.57}{2}}{\quad Time-to-Goal (s)~$\downarrow$} \\
\cmidrule(lr){2-4}
Scenario & GRAM & Robust RL & Contextual RL \\
\midrule
Nominal & \cellcolor[gray]{.9}$\mathbf{7.95 \pm 0.05}$ & $8.07 \pm 0.18$ & $8.62 \pm 0.04$ \\
Frozen Back Hip & \cellcolor[gray]{.9}$\mathbf{7.70 \pm 0.16}$ & $8.79 \pm 0.20$ & $9.33 \pm 0.30$ \\
5\textdegree{} Decline & \cellcolor[gray]{.9}$\mathbf{6.97 \pm 0.17}$ & \cellcolor[gray]{.9}$\mathbf{7.05 \pm 0.16}$ & \cellcolor[gray]{.9}$\mathbf{7.05 \pm 0.15}$ \\
5\textdegree{} Incline & \cellcolor[gray]{.9}$\mathbf{10.03 \pm 0.23}$ & $10.92 \pm 0.50$ & $12.51 \pm 0.35$ \\
\cmidrule(lr){1-4}
Average & \cellcolor[gray]{.9}$\mathbf{8.17}$ \textbf{s} & $8.71$ s & $9.38$ s \\
Success Rate & \cellcolor[gray]{.9}$\mathbf{20/20}$ & \cellcolor[gray]{.9}$\mathbf{20/20}$ & \cellcolor[gray]{.9}$\mathbf{20/20}$ \\
\bottomrule
\end{tabular}
\end{table}



\begin{table}[t]
\footnotesize
\caption{Hardware results: Difficult tasks}
\vspace{-0.25em}
\label{tab:hardware_results_difficult}
\centering
\begin{tabular}{L{0.23} *{3}{C{0.19}}}
\toprule
 & \multicolumn{3}{M{0.57}{2}}{\quad Success Rate~$\uparrow$} \\
\cmidrule(lr){2-4}
Scenario & GRAM & Robust RL & Contextual RL \\
\midrule
9 kg Payload & \cellcolor[gray]{.9}$\mathbf{5/5}$ & \cellcolor[gray]{.9}$\mathbf{5/5}$ & $0/5$ \\
Slippery & \cellcolor[gray]{.9}$\mathbf{5/5}$ & $3/5$ & $2/5$ \\
Ramp to Foam & \cellcolor[gray]{.9}$\mathbf{5/5}$ & $4/5$ & $0/5$ \\
\cmidrule(lr){1-4}
Total & \cellcolor[gray]{.9}$\mathbf{15/15}$ & $12/15$ & $2/15$  \\
\bottomrule
\end{tabular}
\vspace{-1.5em}
\end{table}


In 4 of the 7 scenarios, we considered moderate environment changes where all algorithms successfully reached the goal in all trials. We summarize the results from these scenarios in \tabref{tab:hardware_results_moderate}. GRAM achieves the fastest time-to-goal in each of these scenarios by balancing adaptive and robust behavior at deployment. Robust RL can be overly conservative, which leads to slower time-to-goal in these moderate scenarios. Contextual RL suffers from a sim-to-real gap due to its lack of robustness when only trained on the moderate variations described by $\mathcal{C}_{{\textnormal{ID}}}$ in \tabref{tab:base_id}, resulting in the slowest time-to-goal across all scenarios.

\begin{figure}[t]
    \centering
    \includegraphics[width=1.00\linewidth]{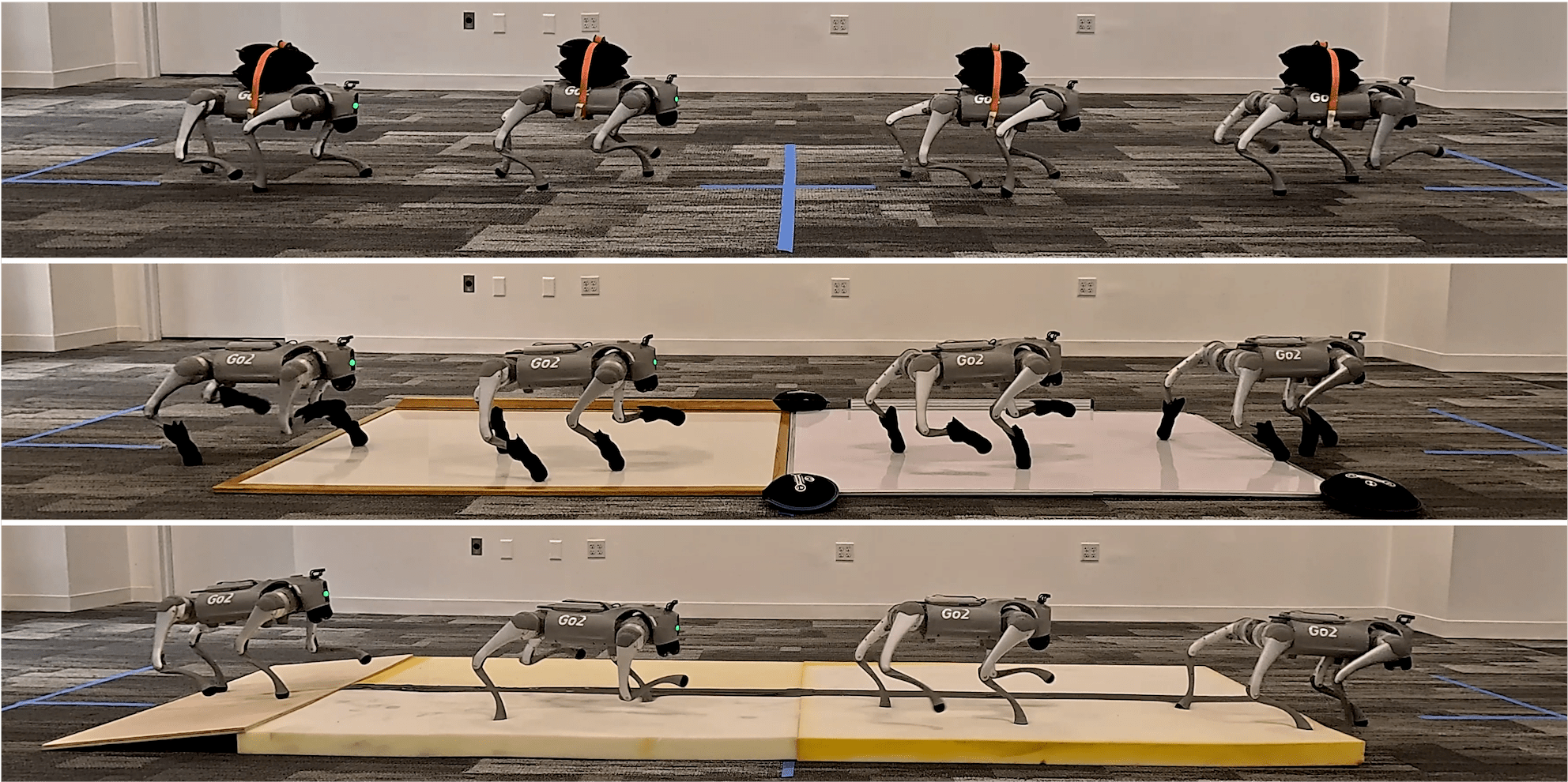}
    \vspace{-1.5em}
    \caption{Difficult controlled hardware experiments where at least one baseline algorithm fails. GRAM achieves 100\% success rate on these tasks. From top to bottom: 9~kg~Payload, Slippery, Ramp to Foam.}
    \label{fig:hardware_indoor}
\end{figure}

In the remaining 3 out of 7 scenarios, we considered more difficult environment variations that differ significantly from those seen during training (see \figref{fig:hardware_indoor} and \tabref{tab:hardware_results_difficult}). Because these scenarios represent far OOD environment dynamics relative to the set of ID training environments, contextual RL only reached the goal in 2 out of 15 trials. Many of these failures occurred when dynamics changed within a trial, such as when the robot first stepped onto the foam or slippery whiteboard. Robust RL successfully reached the goal in 12 out of 15 trials, with failures caused by the robot becoming stuck due to the conservative wide gait learned from robust training. Unlike the baseline algorithms, GRAM achieves 100\% success rate in these difficult scenarios, demonstrating the robust generalization capabilities of our algorithm.


\paragraph{Qualitative results}

\begin{figure}[t]
    \centering
    \includegraphics[width=1.00\linewidth]{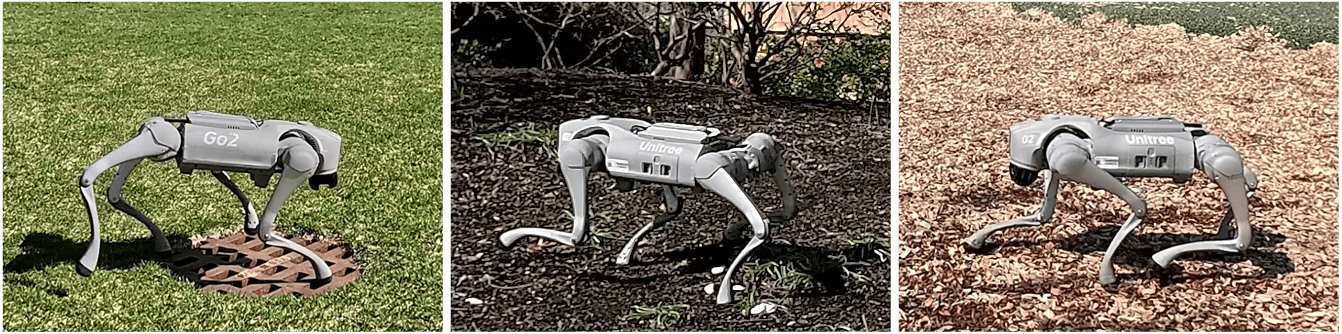}
    \vspace{-1.5em}
    \caption{Qualitative outdoor experiments. GRAM is trained in simulation on flat, solid ground, and achieves robust zero-shot sim-to-real transfer across a variety of outdoor terrains not seen during training. From left to right: grass with storm drain, uneven dirt, wood chips.}
    \label{fig:hardware_outdoor}
    \vspace{-1.0em}
\end{figure}

In addition to controlled trials, we tested the generalization capabilities of GRAM through a variety of qualitative outdoor experiments (see \figref{fig:hardware_outdoor}). Despite only training on flat, solid ground in simulation, GRAM successfully transfers zero-shot to a variety of real-world terrains including pavement, grass, uneven dirt, and wood chips. In addition, GRAM is robust to transitions across terrains, inclines, and other unfamiliar surfaces such as the storm drain shown in \figref{fig:hardware_outdoor}. These experiments further demonstrate GRAM's ability to achieve robust locomotion in both ID and OOD environments at deployment time. See the Supplementary Material for videos of all experiments.


\section{Conclusion}

In this work, we have presented a deep RL framework that achieves both ID and OOD dynamics generalization at deployment time within a single architecture. Our algorithm GRAM leverages a robust adaptation module that allows for adaptation in ID contexts, while also identifying OOD environments with a special robust latent feature $z_{\textnormal{rob}}$. We presented a training pipeline that jointly trains for adaptive ID performance and robust OOD performance, resulting in strong generalization capabilities across a range of simulation and hardware locomotion experiments with a Unitree Go2 quadruped robot. The ability to achieve ID and OOD generalization within a unified framework is critical for the reliable deployment of deep RL in real-world settings, and GRAM represents a principled step towards this goal.


\paragraph{Limitations and future work} 

Because OOD contexts are unknown during training by definition, the OOD generalization of GRAM depends on how well the robust RL training pipeline captures worst-case OOD dynamics, which may lead to failures in extreme OOD scenarios. We applied a standard choice of adversary that worked well in our controlled hardware trials and qualitative outdoor experiments on a quadruped robot, but it would be interesting to combine GRAM with other robust RL techniques and explore its generalization capabilities across a broader range of unstructured environments. Beyond quadruped locomotion, there are also opportunities to apply GRAM across other robotic control applications where generalization is important. For example, GRAM is a flexible deep RL framework that can be extended to other robots and tasks, where domain-specific knowledge can be incorporated through the choice of ID context set for adaptive training and the choice of adversary for robust training.


\bibliographystyle{IEEEtran}
\bibliography{IEEEabrv,bibliography}


\end{document}